\documentclass{article}

\usepackage[numbers]{natbib}
\usepackage[final]{nips_2016}

\usepackage[utf8]{inputenc} 
\usepackage[T1]{fontenc}    
\usepackage{hyperref}       
\usepackage{url}            
\usepackage{booktabs}       
\usepackage{array,booktabs,calc}
\usepackage{amsfonts}       
\usepackage{amsmath}
\usepackage{bm}
\usepackage{nicefrac}       
\usepackage{microtype}      
\usepackage{algorithm}
\usepackage{algorithmic}
\usepackage{color}
\usepackage[dvipsnames]{xcolor}
\usepackage{graphicx}

\DeclareMathOperator*{\argmax}{arg\,max}
\DeclareMathOperator*{\argmin}{arg\,min}

\title{Bayesian Optimization for Machine Learning \\ A Practical Guidebook}

\author{
    Ian Dewancker \hspace{5mm}  Michael McCourt \hspace{5mm} Scott Clark \vspace{3mm} \\
    SigOpt \\
    San Francisco, CA 94108 \\
    \texttt{\{ian, mike, scott\}@sigopt.com} 
}

\begin{document}

\maketitle

\begin{abstract}
  The engineering of machine learning systems is still a nascent field; relying on a seemingly daunting collection of quickly evolving tools and best practices.  
  It is our hope that this guidebook will serve as a useful resource for machine learning practitioners looking to take advantage of Bayesian optimization techniques.  
  We outline four example machine learning problems that can be solved using open source machine learning libraries, and highlight the benefits of using Bayesian optimization 
  in the context of these common machine learning applications.
  
\end{abstract}

\section{Introduction}

Recently, there has been interest in applying Bayesian black-box optimization strategies to better conduct optimization over hyperparameter configurations of machine learning models and systems \cite{snoek2012practical} \cite{ThorntonHutter13} \cite{feurer-nips2015}. 
Most of these techniques require that the objective be a scalar value depending on the hyperparamter configuration $\mathbf{x}$.  
\begin{align*}
\mathbf{x}_{opt} = \argmax_{\mathbf{x} \in \mathcal{X}}  f(\mathbf{x})
\end{align*}
A more detailed introduction to Bayesian optimization and related techniques is provided in \cite{DewanckerBOPrimer2015}.
The focus of this guidebook is on demonstrating several example problems where Bayesian optimization provides a noted benefit.  
Our hope is to clearly show how Bayesian optimization can assist in better
designing and optimizing real-world machine learning systems.  All of the examples in this guidebook
have corresponding code available on SigOpt's example github repo. 

\section{Tuning Text Classification Pipelines with scikit-learn }

Text classification problems appear quite often in modern information systems, and you might imagine building a small document / tweet / blogpost classifier for any number of purposes.  In this example, the classification task is to label Amazon product reviews \cite{blitzer2007biographies} as either favorable or not.  The objective is to find a classifier that is accurate in its predictions, but also one that gives us confidence it will generalize to data it has not been trained on.  We employ the Swiss army knife of machine learning, logistic regression (LR), as our model in this experiment.  While the LR model might be conceptually simple \cite{murphy2012machine} and implemented in many statistics and machine learning software packages, valuable engineering time and resources are often wasted experimenting with feature 
representation and parameter tuning via trial and error.

\subsection{Objective Metric : $f(\pmb{\lambda})$}

SigOpt finds parameter configurations that maximize any metric, so we need to pick one that is appropriate for this classification task. We’ll use $f(\pmb{\lambda})$ to denote our objective metric function and $\pmb{\lambda}$ to represent the set of tunable parameters, which we discuss in the following section.  In designing our objective metric, accuracy, the number of correctly classified reviews, is obviously important, but we also want assurance that our model generalizes and can perform well on data on which it was not trained.  This is where the idea of cross-validation comes into play.

Cross-validation requires us to split up our entire labeled dataset  $\mathcal{D}$ into two distinct sets: one to train on $\mathcal{D}_{\text{train}}$ and one to validate our trained classifier on $\mathcal{D}_{\text{valid}}$.  We then consider metrics like accuracy on only the validation set.  Taking this further and considering not one, but many possible splits of the labeled data is the idea of k-fold cross-validation where multiple training, validation sets are generated and validation metrics can be aggregated in several ways (e.g., mean, min, max) to give a single estimation of performance.

In this case, we’ll use the mean of the k-folded cross-validation accuracies \cite{eggensperger2013towards}.  In our case, $k=5$ folds are used and the train and validation sets are split randomly using 70\% and 30\% of the entire dataset, respectively.
\begin{align*}
\mathcal{L}(\pmb{\lambda}, \mathcal{D}_{\text{t}}, \mathcal{D}_{\text{v}}) &= \text{acc. of LR}(\pmb{\lambda}, \mathcal{D}_{\text{t}}) \text{ on } \mathcal{D}_{\text{v}} \\
f(\pmb{\lambda} ) &= \frac{1}{k} \sum_{i=1}^{k}  \mathcal{L}(\pmb{\lambda}, \mathcal{D}^{(i)}_{\text{train}}, \mathcal{D}^{(i)}_{\text{valid}})
\end{align*}
This objective metric $f(\pmb{\lambda})$ takes on values in the range [0, 1.0], where 0 represents a mis-classification of every example in all validation folds and 1.0 represents perfect classification on all validation folds. The higher the cross-validation metric, the better our classifier is doing.  Using many folds might not be practical if training takes an very long time (you might have to settle for 1 or 2 folds only).

\subsection{Tunable Parameters : $\pmb{\lambda}$}

The objective metric, $f(\pmb{\lambda})$, is controlled by a set of parameters, $\pmb{\lambda}$, that potentially influence its performance.  Parameters can be defined on continuous, integer or categorical domains. The parameters used in this experiment can be split into two groups: those governing the feature representation of the review text and those governing the cost function of logistic regression. We explain these sets of parameters in the following sections. 

\subsubsection{Feature Representation Parameters}

The CountVectorizer class in scikit-learn is a convenient mechanism for transforming a corpus of text documents into vectors using bag of words representations (BOW).  scikit-learn offers quite a bit of control in determining which n-grams make up the vocabulary for your BOW vectors.  As a quick refresher, n-grams are sequences of text tokens as shown below:

\bgroup
\def\arraystretch{1.6}
\begin{table}[H]
	\begin{center}
		\begin{tabular}{ |>{\centering}m{2.5cm}|>{\centering}m{9.5cm}|} 
			\hline
		    \bf{Original Text}  & "SigOpt optimizes any complicated system"   \tabularnewline
			\hline
			\bf{1-grams}  & \{"SigOpt", "optimizes", "any", "complicated", "system" \}   \tabularnewline
			\hline
			\bf{2-grams}  & \{"SigOpt\_optimizes", "optimizes\_any", "any\_complicated" \dots \}   \tabularnewline
			\hline
			\bf{3-grams}  & \{  "SigOpt\_optimizes\_any", "optimizes\_any\_complicated" \dots \}   \tabularnewline
			\hline
		\end{tabular}
		\vspace{4mm}
		\caption{Example n-grams for a sample piece of text}
	\end{center}
\end{table}
\egroup

The number of times each n-gram appears in a given piece of text is then encoded in the BOW vector describing that text.  CountVectorizer allows you to control the range of n-grams that are included in the vocabulary $(min\_n\_gram, ngram\_offset$ in our experiment), as well as filtering n-grams outside a specified document-frequency range $(log\_min\_df, df\_offset$ in our experiment). For example, if a rare 3-gram like "hi\_diddly\_ho" doesn’t appear with at least min-df frequency in the corpus, it is not included in the vocabulary.  Similarly, n-grams that occur in nearly every document (1-grams like "the", "a" etc) can also be filtered using the max-df parameter.  Often when the range of the parameter is very large or very small, it makes sense to look at the parameter on the log scale, as we do with the $log\_min\_df$ parameter.

\subsubsection{Logistic Regression Error Cost Parameters}

Using the SGDClassifier class in scikit-learn, we can succinctly formulate and solve the logistic regression learning problem.  The error function for logistic regression, two-class classification is defined in the following way:

\begin{equation*}
E(\pmb{\theta}) = \frac{1}{M}\sum_{i=1}^{M}\log\left( 1.0 + e^{-y_i(\pmb{\theta}^T\mathbf{x_i}) }\right ) + \alpha\left( \frac{1-\rho}{2}\|\pmb{\theta}\|_2^{2} + \rho\|\pmb{\theta}\|_1\right) \\
\end{equation*}
\vspace{0.85mm}
\begin{align*}
M &= \text{number of training examples} \\
\pmb{\theta} &= \text{vector of weights the algorithm will learn for each n-gram in vocabulary } \\
y_i &= \text{training data label : \{-1, 1\} for our two class problem} \\
\mathbf{x}_i &= \text{training data input vector:  BOW vectors described in previous section} \\
\alpha &= \text{weight of regularization term } \\
\rho &= \text{weight of L1 norm term }
\end{align*}

The first term of the cost function penalizes weights that do not fit the training data while the second term penalizes model complexity (how far are the feature weights away from zero).  scikit-learn performs stochastic gradient descent on this error function with respect to the weights in an attempt to find those that minimize this function.

Should we use L1 or L2 regularization, or perhaps a weighted mixture?  How much should the entire regularization term be weighted?  With this error formulation, and the $\alpha$ and $\rho$ parameters exposed in our experiment, SigOpt can quickly find these answers to these important questions.  

\subsection{Experimental Results}

SigOpt offers one solution to the hyperparameter optimization problem, however there are other existing techniques.  In particular, random search and grid search are two commonly employed strategies.  Random search, as you might guess, simply selects parameter configurations at random, while grid search sweeps through a selected subset of the parameter space.  

How should we evaluate the performance of these alternative optimization strategies?  One criterion that makes sense is to consider the best found (max) value of the objective metric after optimization is complete.  Better performing strategies will find better configurations over the duration of their search.  Due to the stochastic nature of these systems however, we must consider the variation in our best found measurements over several runs to make fair comparisons.  

To ground our discussion, we also report the performance when no hyperparameter optimization is performed, and we simply take the default values for CountVectorizer and SGDClassifier as provided by scikit-learn.  For grid search, we consider 64 evenly spaced parameter configurations (order shuffled randomly) across our domain and analyze the best seen after 60 evaluations to be consistent with our limit on the total number of evaluations for this experiment. Exhaustive grid search is usually prohibitive because the number of possible configurations grows exponentially.
\bgroup
\def\arraystretch{1.6}
\begin{table}[H]
	\begin{center}
		\begin{tabular}{ |>{\centering}m{1.7cm}|>{\centering}m{2.5cm}|>{\centering}m{2cm} |>{\centering}m{1.8cm}|>{\centering}m{2.0cm} | } 
			\hline
			& SigOpt  & Rnd. Search  &  Grid Search & \hspace{0.5mm} No Tuning \newline  (Baseline) \tabularnewline
			\hline
			Best Found \newline ACC & \bf{0.8760} ({\color{ForestGreen}{+5.72\%}}) & 0.8673 & 0.8680 & 0.8286 \tabularnewline
			\hline
		\end{tabular}
		\vspace{4mm}
		\caption{Best found accuracy results averaged over 20 optimization runs, each run consisting of 60 function evaluations}
	\end{center}
\end{table}
\egroup
\vspace{-3mm}
SigOpt finds the best configuration with statistical significance over the other two approaches (p = 0.0001, using the unpaired Mann-Whitney U test) and improves the performance as compared to the baseline by 5.72\%.  

\newpage

\section{Unsupervised Feature Learning with scikit-image and xgboost}

As the previous section discussed, fully supervised learning algorithms require each data point to have an associated class or output. In practice, however, it is often the case that relatively few labels are available during training time and labels are costly or time consuming to acquire. For example, it might be a very slow and expensive process for a group of experts to manually investigate and classify thousands of credit card transaction records as fraudulent or legitimate. A better strategy might be to study the large collection of transaction data without labels, building a representation that better captures the variations in the transaction data automatically.

\subsection{Unsupervised Learning}

Unsupervised learning algorithms are designed with the hope of capturing some useful latent structure in data. These techniques can often enable dramatic gains in performance on subsequent supervised learning task, without requiring more labels from experts. In this post we will use an unsupervised method on an image recognition task posed by researchers at Stanford \cite{coates2010analysis} where we try to recognize house numbers from images collected using Google street view (SVHN). This is a more challenging problem than MNIST (another popular digit recognition data set) as the appearance of each house number varies quite a bit and the images are often cluttered with neighboring digits:
\vspace{-3mm}
\begin{figure}[H]
	\centering
	\includegraphics[width=0.5\linewidth]{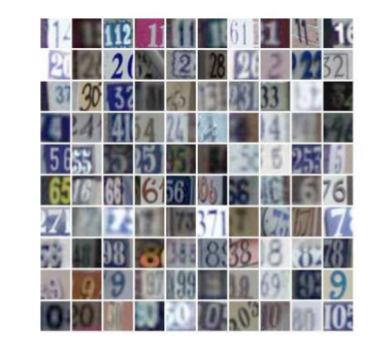}
	\caption{$32\times32$ cropped samples from the classification task of the SVHN dataset. Each sample is assigned only a single digit label (0 to 9) corresponding to the center digit. (Sermanet \cite{sermanet2012convolutional})}
	\label{fig:digits_svhn}
\end{figure}
\vspace{-2mm}
In this example, we assume access to a large collection of unlabelled images $\mathsf{X}_u$, where the correct answer is not known, and a relatively small amount of labelled data $(\mathsf{X}_s, \mathbf{y})$ for which the true digit in each image is known (often requiring a non-trivial amount of time and money to collect). Our hope is to find a suitable unsupervised model, built using our large collection of unlabelled images, that transforms images into a more useful representation for our classification task. 

Unsupervised and supervised learning algorithms are typically governed by small sets of hyperparameters $(\pmb{\lambda}_u, \pmb{\lambda}_s)$, that control algorithm behavior. In our example pipeline below, $\mathsf{X}_u$ is used to build the unsupervised model $f_u$ which is then used to transform the labelled data $(\mathsf{X}_s, \mathbf{y})$ before the supervised model $f_s$ is trained. Our task is to efficiently search for good hyperparameter configurations $(\pmb{\lambda}_u, \pmb{\lambda}_s)$ for both the unsupervised and supervised algorithms. SigOpt minimizes the classification error $E(\pmb{\lambda}_u, \pmb{\lambda}_s)$ by sequentially generating suggestions for the hyperparameters of the model $(\pmb{\lambda}_u, \pmb{\lambda}_s)$. For each suggested hyperparameter configuration a new unsupervised data representation is formed and fed into the supervised model. The observed classification error is reported and the process repeats, converging on the set of hyperparameters that minimizes the classification error.
\begin{figure}[H]
	\centering
	\includegraphics[width=\linewidth]{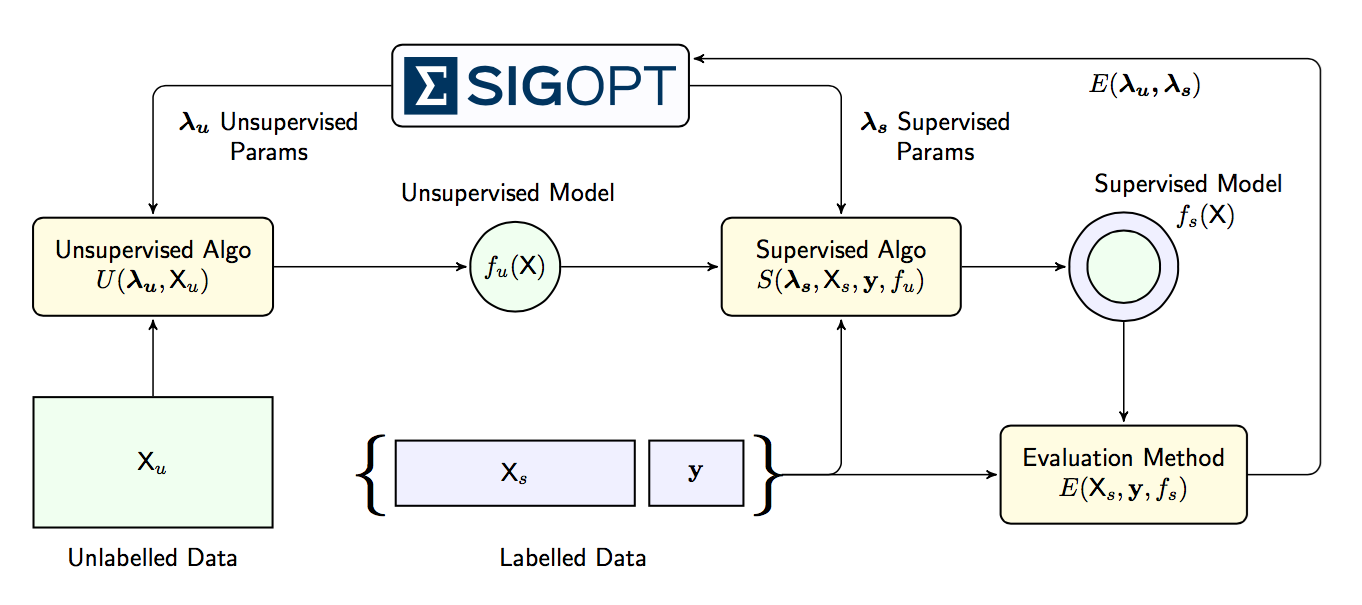}
	\caption{Process for coupled unsupervised and supervised model tuning.}
	\label{fig:ml_pipe}
\end{figure}

SigOpt offers Bayesian optimization as a service, capable of efficiently searching through the joint variations $(\pmb{\lambda}_u, \pmb{\lambda}_s)$ of both the supervised and unsupervised aspects of machine learning systems, as depicted in Figure \ref{fig:ml_pipe}.  This allows experts to unlock the power of unsupervised strategies with the assurance that each model is reaching its full potential automatically.  

\subsection{Unsupervised Model}

We start with the initial features describing the data: raw pixel intensities for each image. The goal of the unsupervised model is to transform the data from its original representation to a new (more useful) learned representation without using labeled data. Specifically, you can think of this unsupervised model as a function $f : \mathbb{R}^N \to \mathbb{R}^J$. Where $N$ is the number of features in our original representation and $J$ is the number of features in the learned representation. In practice, expanded representations (sometimes referred to as a feature map) where $J$ is much larger than $N$ often work well for improving performance on classification tasks \cite{bengio2012deep}.

\subsubsection{Image Transform Parameters ($s, w, K$)}

A simple but surprisingly effective transformation for small images was proposed in a paper by Coates \cite{coates2010analysis} where image patches are transformed into distances to $K$ learned centroids (average patches) using the k-means algorithm, and then pooled together to form a final feature representation as outlined in Figure \ref{fig:coates_transform} below:
\begin{figure}[H]
	\centering
	\includegraphics[width=0.85\linewidth]{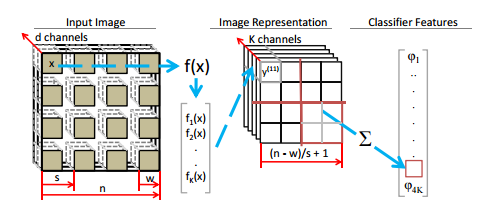}
	\caption{Feature extraction using a $w \times w$ receptive field and stride $s$. $w \times w$ patches separated by $s$ pixels each, then map them to $K$-dimensional feature vectors to form a new image representation. The vectors are then pooled over the image quadrants to form the classifier feature vector. Coates \cite{coates2010analysis}}
	\label{fig:coates_transform}
\end{figure}

In this example we are working with the 32x32 (n=32) converted gray-scale (d=1) images of the SVHN dataset. We allow SigOpt to vary the stride length ($s$) and patch width ($w$) parameters. The figure above illustrates a pooling strategy that considers quadrants in the 2x2 grid of the transformed image representation, summing them to get the final transformed vector. We used the suggested resolution in \cite{coates2010analysis} and kept $pool_r$ fixed at 2. $\mathsf{f}(x)$ represents a $K$ dimensional vector that encodes the distances to the $K$ learned centroids, and $\mathsf{f}_i(x)$ refers to the distance of image patch instance $x$ to centroid $i$. In this experiment, $K$ is also a tunable parameter. The final feature representation of each image will have $J = K \cdot pool_r^2$ features.

\subsubsection{Whitening Transform Parameter ($\epsilon_{\text{zca}}$)}

Before generating the image patch centroids and any subsequent patch comparisons to these centroids, we apply a whitening transform to each patch. When dealing with image data, whitening is a common preprocessing transform which removes the correlation between all pairs of individual pixels \cite{krizhevsky2009learning}. Intuitively, it can be thought of as a transformation that highlights contrast in images. It has been shown to be helpful in image recognition tasks, and may also be useful for other feature data. The figure below shows several example image patches before and after the whitening transform.
\vspace{-3mm}
\begin{figure}[H]
	\centering
	\includegraphics[width=0.75\linewidth]{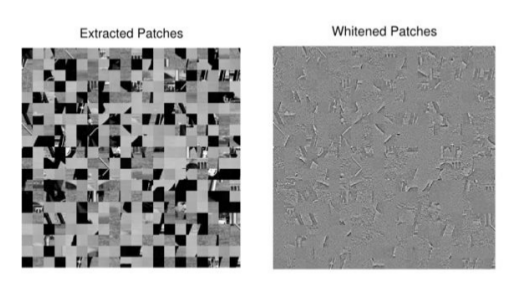}
	\caption{Comparison of image patches before and after whitening ( Stansbury \cite{Stansbury:2014} )}
	\label{fig:whitening}
\end{figure}
The whitening transformation we use is known as ZCA whitening \cite{coates2012learning}. This transform is achieved by cleverly applying the eigendecomposition of the covariance matrix estimate to a mean adjusted version of the data matrix, so that the expected covariance of the data matrix becomes the identity. A regularization term $\epsilon_{\text{zca}}$ is added to the diagonal eigenvalue matrix, and $\epsilon_{\text{zca}}$ is exposed as a tunable parameter to SigOpt.
\begin{align*}
\text{cov}(\mathsf{X}) &= \mathsf{U \Lambda U^T} \\
\mathsf{\Lambda}^{-\frac{1}{2}} &= \text{diag}(1/ \sqrt{\mathsf{\Lambda}_{ii}}) \\
\mathsf{X}_{\text{zca}} &= (\mathsf{X} - \mathbf{1}\boldsymbol{\mu}^\mathsf{T}) \mathsf{U} (\mathsf{\Lambda}+ \epsilon_{\text{zca}}\mathsf{I})^{-\frac{1}{2}} \mathsf{U^T}
\end{align*}

\subsubsection{Centroid Distance Sparsity Parameter ($sparse_p$)}

Each whitened patch in the image is transformed by considering the distances to the learned $K$ centroids. To control this sparsity of the representation we report only distances that are below a certain percentile, $sparse_p$, when considering the pairwise distances between the current patch and the centroids. Intuitively this acts as a threshold which allows for only the “close” centroids to be active in our representation.

Figure \ref{fig:sparse_cent} below illustrates the idea with a simplified example. A whitened image patch (in the upper right) is compared against the 4 learned centroids after k-means clustering. Here, let’s imagine we have set the percentile threshold to 50, so only the distances in the lower half of all centroid distances persist in the final representation, the others are zeroed out

\begin{figure}[H]
	\centering
	\includegraphics[width=0.7\linewidth]{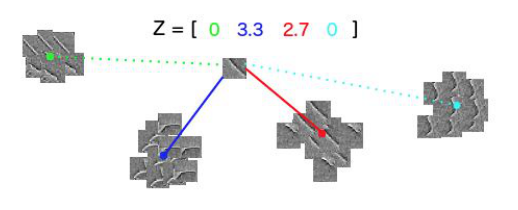}
	\caption{Sparsity transform; distances from a test patch to centroids $>$ 50th percentile are set to 0}
	\label{fig:sparse_cent}
\end{figure}

While the convolutional aspects of this unsupervised model are tailored to image data, the general approach of transforming feature data into a representation that reflects distances to learned archetypes seems suitable for other data sets and feature spaces \cite{dieleman2013multiscale}.

\subsection{Supervised Model}

With the learned representation of our data, we now seek to maximize performance on our classification task using a smaller labelled dataset. While random forests are an excellent, and simple, classification tool, better performance can typically be achieved by using carefully tuned ensembles of boosted classification trees. 

\subsubsection{Gradient Boosting Parameters ($\gamma, \theta, M$)}

We consider the popular library XGBoost as our gradient boosting implementation. Gradient boosting is a generic boosting algorithm that incrementally builds an additive model of base learners, which are themselves simpler classification or regression models. Gradient boosting works by building a new model at each iteration that best reconstructs the gradient of the loss function with respect to the previous ensemble model. In this way it can be seen as a sort of functional gradient descent, and is outlined in more detail below. In the pseudocode below we outline building an ensemble of regression trees, but the same method can be used with a classification loss function $L$

\begin{algorithm}[H]
	\caption{Gradient Boost \label{alg:gradboost}}
	\begin{algorithmic}
		\STATE {\bfseries Input:} $\mathcal{D} = \{ (\mathbf{x}_1, y_1 ),\, \ldots,\, (\mathbf{x}_N, y_N) \}, \theta, \gamma$
		\STATE {\bfseries Output:} $F(\mathbf{x}) \ = \sum_{i=0}^{M} F_i(\mathbf{x})$
		\STATE $F_0(\mathbf{x}) \gets \argmin_\beta \sum_{i=1}^N L(y_i, \beta)$
		\FOR{$m \gets 1$ {\bfseries to} $M$}
		\vspace{1.5mm}
		\STATE $d_i = -\left[ \frac{\partial L(y_i, F(\mathbf{x}_i))}{\partial F(\mathbf{x}_i) } \right]_{F(\mathbf{x}_i) = F_{m-1}(\mathbf{x}_i)}$
		\vspace{1mm}
		\STATE $ \mathcal{G} \gets \left \{ \left ( \mathbf{x}_i, d_i \right )\right \}, i = 1,N $
		\STATE $ g(\mathbf{x}) \gets \textsc{FitRegrTree}(\mathcal{G}, \theta) $
		\STATE $\rho_m \gets \argmin_{\rho} \sum_{i=1}^N L(y_i, F_{m-1}(\mathbf{x}) + \rho g(\mathbf{x})) $
		\STATE $ F_{m}(\mathbf{x}) \gets F_{m-1}(\mathbf{x}) + \gamma \ \rho_mg(\mathbf{x})$ 
		\ENDFOR
	\end{algorithmic}
\end{algorithm}

\subsection{Experimental Results}

We compare the ability of SigOpt to find the best hyperparameter configuration to random search, which usually outperforms grid search and manual search (Bergstra \cite{bergstra2012random}) and a baseline of using an untuned model.

Because the underlying methods used are inherently stochastic we performed 10 independent hyperparameter optimizations using both SigOpt and random search for both the purely supervised and combined models. Hyperparameter optimization was performed on the accuracy estimate from a 80/20 cross validation fold of the training data (73k examples). The ‘extra’ set associated with the SVHN dataset (530K examples) was used to simulate the unlabelled data $\mathsf{X}_u$ in the unsupervised parts of this example.

For the unsupervised model 90 sequential configuration evaluations (~50 CPU hrs) were used for both SigOpt and random search. For the purely supervised model 40 sequential configuration evaluations (~8 CPU hrs) were used for both SigOpt and random search. In practice, SigOpt is usually able to find good hyperparameter configurations with a number of evaluations equal to 10 times the number of parameters being tuned (9 for the combined model, 4 for the purely supervised model). The same parameters and domains were used for XGBoost in both the unsupervised and purely supervised settings. As a baseline, the hold out accuracy of an untuned scikit-learn random forest using the raw pixel intensity features.

After hyperparameter optimization was completed for each method we compared accuracy using a completely held out data set (SHVN test set, 26k examples) using the best configuration found in the tuning phase. The hold out dataset was run 10 times for each best hyperparameter configuration for each method, the mean of these runs is reported in the table below. SigOpt outperforms random search with a p-value of 0.0008 using the unpaired Mann-Whitney U test.

\bgroup
\def\arraystretch{1.6}
\begin{table}[H]
	\begin{center}
		\begin{tabular}{ |>{\centering}m{1.4cm}|>{\centering}m{2.5cm}|>{\centering}m{2.0cm} |>{\centering}m{1.9cm}|>{\centering}m{1.9cm} |>{\centering}m{1.9cm} | } 
			\hline
			& \hspace{4mm} SigOpt \newline (xgboost + \newline Unsup. Feats)  & Rnd Search \newline (xgboost + \newline Unsup. Feats) &  SigOpt \newline (xgboost + \newline Raw Feats) &  Rnd Search \newline (xgboost + \newline Raw Feats) &  No Tuning \newline (sklearn RF + \newline Raw Feats) \tabularnewline
			\hline
			Hold out \newline ACC & \bf{0.8601} ({\color{ForestGreen}{+49.2\%}}) & 0.8190 & 0.7483 & 0.7386 & 0.5756 \tabularnewline
			\hline
		\end{tabular}
		\vspace{4mm}
		\caption{Comparison of model accuracy on held out (test) dataset after different tuning strategies}
	\end{center}
\end{table}
\egroup

The chart below in Figure \ref{fig:unsup_trace} shows the optimization traces of SigOpt versus random search optimization strategies when tuning the unsupervised model (Unsup Feats) and only the supervised model (Raw Feats). We plot the interquartile range of the best seen cross validated accuracy score on the training set at each objective evaluation during the optimization. As mentioned above, 90 objective evaluations were used in the optimization of the unsupervised model and 40 in the supervised setting. SigOpt outperforms random search in both settings on this training data (p-value 0.005 using the same Mann-Whitney U test as before). 

\begin{figure}[H]
	\centering
	\includegraphics[width=\linewidth]{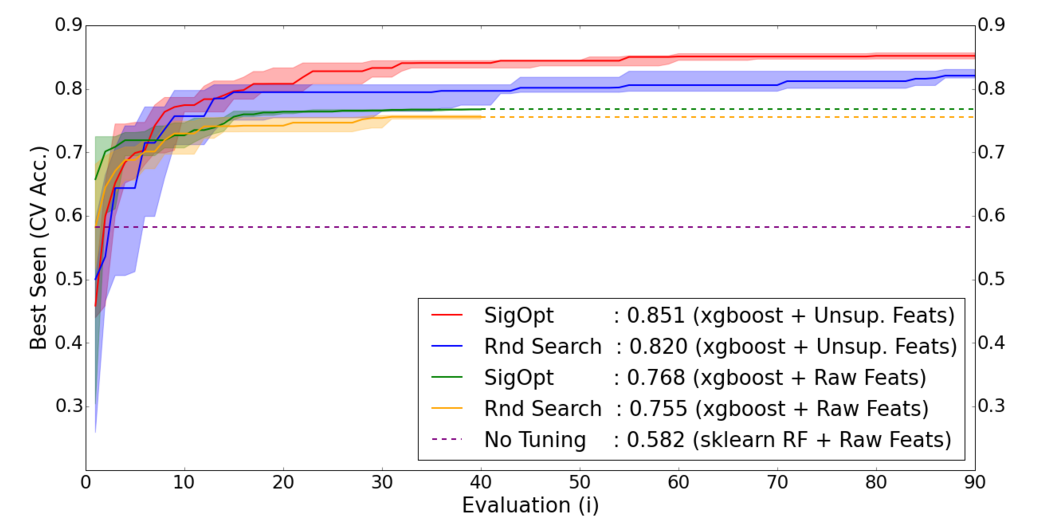}
	\caption{Optimization traces of CV accuracy using SigOpt and random search.}
	\label{fig:unsup_trace}
\end{figure}

\newpage

\section{Deep Learning with TensorFlow}

There are a large number of tunable parameters associated with defining and training deep neural networks \cite{bengio2009learning} \cite{bergstra2011algorithms} and SigOpt accelerates searching through these settings to find optimal configurations. This search is typically a slow and expensive process, especially when using standard techniques like grid or random search, as evaluating each configuration can take multiple hours. SigOpt finds good combinations far more efficiently than these standard methods by employing an ensemble of Bayesian optimization techniques.

In this example, we consider the same optical character recognition task of the SVHN dataset as discussed in the previous section. Our goal is to build a model capable of recognizing digits (0-9) in small, real-world images of house numbers. We use SigOpt to efficiently find a good structure and training configuration for a convolutional neural net.

\subsection{Convolutional Neural Net Structure}

\begin{figure}[H]
	\centering
	\includegraphics[width=\linewidth]{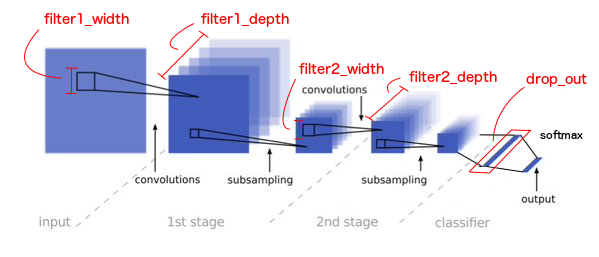}
	\caption{Representative convolutional neural net topology. Important parameters include the width and depth of the convolutional filters, as well as dropout probability \cite{sermanet2012convolutional}}
	\label{fig:cnn_example}
\end{figure}

The structure and topology of a deep neural network can have dramatic implications for performance on a given task \cite{bengio2009learning}. Many small decisions go into the connectivity and aggregation strategies for each of the layers that make up a deep neural net. These parameters can be non-intuitive to choose in an optimal, or even acceptable, fashion. In this experiment we used a TensorFlow CNN example designed for the MNIST dataset as a starting point. Figure \ref{fig:cnn_example} represents a typical CNN structure, highlighting the parameters we chose to vary in this experiment. A more complete discussion of these architectural decisions can be found in an online course from Stanford ( Li \cite{Li:2015} ). It should be noted that Figure \ref{fig:cnn_example} is an approximation of the architecture used in this example, and the code in the SigOpt examples repository serves as a more complete reference. 
\vspace{-2mm}
TensorFlow has greatly simplified the effort required to build and experiment with deep neural network (DNN) designs. Tuning these networks, however, is still an incredibly important part of creating a successful model. The optimal structural parameters often highly depend on the dataset under consideration.

\subsection{Stochastic Gradient Descent Parameters ($\alpha, \beta, \gamma$)}

\begin{figure}[H]
	\centering
	\includegraphics[width=\linewidth]{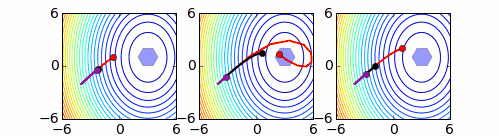}
	\caption{Progression of RMSProp gradient descent after 12 update steps under different parametrizations. left: Various decay rates with other parameters fixed: purple = .01, black = .5, red = .93. center: Various learning rates with other parameters fixed: purple = .016, black = .1, red = .6. right: Various momentums with other parameters fixed: purple = .2, black = .6, red = .93.}
	\label{fig:sgd_example}
\end{figure}

Once the structure of the neural net has been selected, an optimization strategy based on stochastic gradient descent (SGD) is used to fit the weight parameters of the convolutional neural net. There is no shortage of SGD algorithm variations implemented in TensorFlow.  To demonstrate how drastically their behavior can vary under different parameterizations, Figure \ref{fig:sgd_example} compares several configurations of RMSProp, a particular SGD variation on a simple 2D objective.

It can be a counterintuitive and time consuming task to optimally configure a particular SGD algorithm for a given model and dataset. To simplify this tedious process, we expose to SigOpt the parameters that govern the RMSProp optimization algorithm. Important parameters governing its behavior are the learning rate $\alpha$ , momentum $\beta$ and decay $\gamma$ terms. These parameters define the RMSProp gradient update step, outlined in the pseudo code below:

\begin{algorithm}[H]
	\caption{RMSProp Stochastic Gradient Descent \label{alg:sgdalgorithm}}
	\begin{algorithmic}
		\STATE {\bfseries Input:}  $\nabla_{\pmb{\theta}}f(\pmb{\theta}) $, $\pmb{\theta}_0$, $\alpha$, $\beta$, $\gamma$, $\epsilon$
		\STATE $\mathbf{m}_0  \gets \mathbf{0}$
		\STATE $\mathbf{b}_0  \gets \mathbf{0}$
		\FOR{$t \gets 1$ {\bfseries to} $T$}
		\STATE $\mathbf{g} \gets \nabla_{\pmb{\theta}}f(\pmb{\theta}_{t-1})$  \hspace{3mm} stochastic gradient 
		\STATE $\mathbf{m}_t[i] \gets \gamma \mathbf{m}_{t-1}[i] + (1-\gamma)\mathbf{g}[i]^2 \hspace{12.5mm} i = 1\dots N$
		\STATE $\mathbf{b}_t[i] \gets \beta \mathbf{b}_{t-1}[i] + \alpha\left( \frac{\mathbf{g}[i]}{ \sqrt{(\mathbf{m}_t[i]+ \epsilon)}} \right) \hspace{8mm} i = 1\dots N$
		\STATE $\pmb{\theta}_t \gets \pmb{\theta}_{t-1} - \mathbf{b}$
		\ENDFOR
	\end{algorithmic}
\end{algorithm}

For this example, we used only a single epoch of the training data, where one epoch refers to a complete presentation of the entire training data (~500K images in our example). Batch size refers to the number of training examples used in the computation of each stochastic gradient (10K images in our example). One epoch is made up of several batch sized updates, so as to minimize the in-memory resources associated required for the optimization (Hinton \cite{Hinton:2015}). Using only a single epoch can be detrimental to performance, but this was done in the interest of time for this example.

\subsection{Experimental Results}

To compare tuning the CNNs hyperparameters when using random search versus SigOpt, we ran 5 experiments using each method and compared the median best seen trace. The objective was the classification accuracy on a single 80 / 20 fold of the training and "extra" set of the SVHN dataset (71K + 500K images respectively). The median best seen trace for each optimization strategy is shown below in Figure \ref{fig:conv_plot}.

In our experiment we allowed SigOpt and random search to perform 80 function evaluations (each representing a different proposed configuration of the CNN). A progression of the best seen objective at each evaluation for both methods is shown below in Figure \ref{fig:conv_plot}. We include, as a baseline, the accuracy of an untuned TensorFlow CNN using the default parameters suggested in the official TensorFlow example. We also include the performance of a random forest classifier using sklearn defaults. 

\begin{figure}[H]
	\centering
	\includegraphics[width=\linewidth]{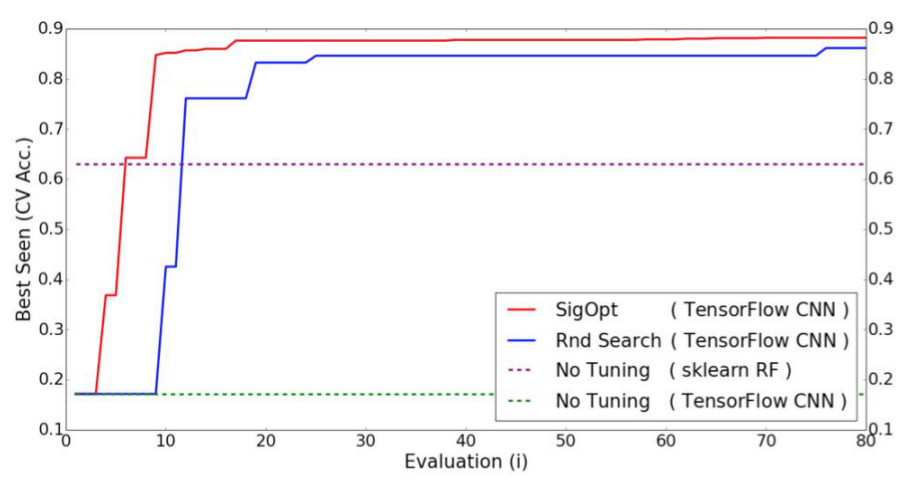}
	\caption{Median best seen trace of CV accuracy over 5 independent optimization runs using SigOpt, random search as well as two baselines where no tuning was performed.}
	\label{fig:conv_plot}
\end{figure}

After hyperparameter optimization was completed for each method, we compared accuracy using a completely held out data set (SHVN test set, 26K images) using the best configuration found in the tuning phase.  The best hyperparameter configurations for each method in each of the 5 optimization runs was used for evaluation. The mean of these accuracies is reported in the table below. We also include the same baseline models described above and report their performance on the held out evaluation set.

\bgroup
\def\arraystretch{1.6}
\begin{table}[H]
	\begin{center}
		\begin{tabular}{ |>{\centering}m{1.25cm}|>{\centering}m{2.9cm}|>{\centering}m{3cm} |>{\centering}m{1.8cm}|>{\centering}m{2.9cm} | } 
			\hline
			& \hspace{4mm} SigOpt \newline (TensorFlow CNN)  & Random Search \newline  (TensorFlow CNN) &  No Tuning \newline (sklearn RF) & \hspace{4mm} No Tuning \newline  (TensorFlow CNN) \tabularnewline
			\hline
			Hold out \newline ACC & \bf{0.8130} ({\color{ForestGreen}{+315.2\%}}) & 0.5690 & 0.5278 & 0.1958 \tabularnewline
			\hline
		\end{tabular}
		\vspace{4mm}
		\caption{Comparison of model accuracy on the held out (test) dataset after different tuning strategies}
	\end{center}
\end{table}
\egroup

\newpage

\section{Recommendation Systems with MLlib}

A popular approach for building the basis of a recommendation system is to learn a model capable of predicting users' product preferences or ratings.  With an effective predictive model, and enough contextual information about users, online systems can better suggest content or products, helping to promote  sales, subscriptions or conversions.  
\vspace{-4mm}
\begin{figure}[H]
	\centering
	\includegraphics[width=\linewidth]{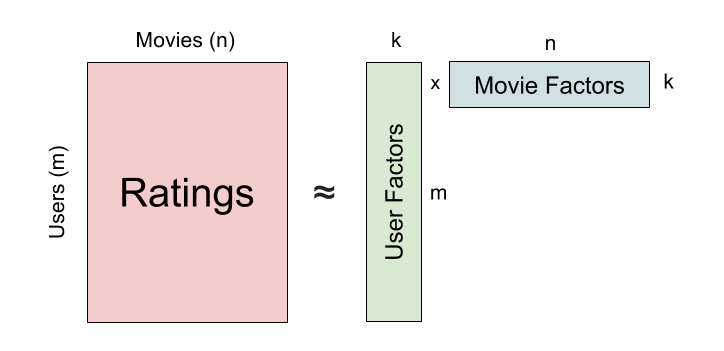}
	\caption{Collaborative Filtering via Low-Rank Matrix Factorization}
\end{figure}
A common recommender systems model involves using a low-rank factorization of a user-product ratings matrix to predict the ratings of other products for each user \cite{koren2008factorization}.  In general, algorithms related to collaborative filtering and recommendation systems will have tunable parameters similar to ones we have discussed in previous sections.  In this problem, for example, the regularization term on the user and product factors can be difficult to choose a priori without some trial and error. 

In this example we consider the MovieLens dataset and use the MLlib package within Apache Spark. The code for this example is available in the SigOpt examples github repository.  We use the largest MovieLens dataset ratings matrix which has approximately 22 million user ratings for 33,000 movies by 240,000 users.  To run this example, we recommend creating a small spark cluster in ec2 using the spark-ec2 tool provided in the spark library.  We ran this experiment using a 3 machine cluster (1 master, 2 workers) in AWS using the m1.large instance for all nodes.

\subsection{Alternating Least Squares}

To solve for the latent user and movie factors, MLlib implements a variant of what is known as quadratically regularized PCA \cite{udell2016generalized}.  Intuitively, this optimization problem aims to learn latent factors $X,Y$ that best recreate the ratings matrix $A$, with a regularization penalty coefficient $\lambda$ on the learned factors.  Here $\mathbf{x}_i$ represents the $i$th row of the $X$ factor matrix and $\mathbf{y}_j$ represents the $j$th column of the $Y$ factor matrix.
\begin{equation*}
\argmin_{\mathbf{x}_i, \mathbf{y}_j} \sum_{i=1}^{m} \sum_{j=1}^{n} ( A_{ij} - \mathbf{x}_i\mathbf{y}_j)^2 + 
\lambda \sum_{i=1}^{m}||\mathbf{x}_i||_2^{2} + \lambda \sum_{j=1}^{n}||\mathbf{y}_j||_{2}^{2}
\end{equation*}

This minimization problem can be solved using a technique known as alternating least squares \cite{udell2016generalized} . A distinct advantage of using this formulation is that it can be easily parallelized into many independent least square problems as outlined in the pseudocode below.  Each factor matrix $X,Y$ is randomly initialized and the algorithm alternates between solving for the user factors $X$, holding the movie factors $Y$ constant, then solving for the $Y$ factors, holding $X$ constant.  The algorithm takes as input $A$ the ratings matrix, $\lambda$ the regularization term, $k$ the desired rank of the factorization, and $T$ the number of iterations of each alternating step in the minimization.  We expose $\lambda$, $k$ and $T$ as tunable parameters to SigOpt.

\begin{algorithm}[H]
	\caption{Parallel Alternating Least Squares\label{alg:smboalgorithm}}
	\begin{algorithmic}
		\STATE {\bfseries Input:}  $A \in \mathbb{R}^{m \times n} $, $\lambda$, $k$, $T$
		\STATE $X  \gets \textsc{RandInit}( m, k ) \hskip2em \triangleright \text{ Initialize factors }$
		\STATE $Y  \gets \textsc{RandInit}( k, n )$
		\FOR{$iter \gets 1$ {\bfseries to} $T$}
		\vspace{1mm}
		\STATE \textbf{par for}\hspace{1mm}$i \gets 1$ {\bfseries to} $m \hskip2em\triangleright \text{ Executed in parallel  }$
		\STATE $ \hskip2em \mathbf{x}_{i} \gets \argmin\limits_{ \mathbf{x}_{i}} || \mathbf{x}_{i} Y - A_{i,*}  ||_2^2 + \lambda ||\mathbf{x}_{i}||_2^2$
		\vspace{1mm}
		\STATE \textbf{par for}\hspace{1mm}$j \gets 1$ {\bfseries to} $n \hskip2em\triangleright \text{ Executed in parallel  }$
		\STATE $\hskip2em \mathbf{y}_{j} \gets  \argmin\limits_{\mathbf{y}_{j}}  || X\mathbf{y}_{j} - A_{*,j}  ||_2^2 + \lambda ||\mathbf{y}_{j}||_2^2 $
		\vspace{1mm}
		\ENDFOR
	\end{algorithmic}
\end{algorithm}

The regularization term $\lambda$ is particularly difficult to select optimally as it can drastically change the generalization performance of the algorithm. Previous work has attempted to use a Bayesian formulation of this problem to avoid optimizing for this regularization term explicitly \cite{salakhutdinov2007probabilistic}

\subsection{Experimental Results}

As an error metric for this example, we used the standard measurement of the root mean square error \cite{koren2008factorization} of the reconstructions on a random subset of nonzero entries from the ratings matrix.
\vspace{1mm}
\begin{equation*}
\text{RMSE} = \sqrt{ \sum_{(i,j) \in TestSet} \frac{(A_{ij} - \mathbf{x}_i \mathbf{y}_j)^2}{|TestSet|}}
\end{equation*}

Defining an appropriate error measurement for a recommendation task is critical for achieving success.  Many other metrics have been proposed for evaluating recommendation systems and careful selection is required to tune for models that are best for the application at hand. Bayesian optimization methods like SigOpt can be used to tune any underlying metric, or a composite metric of many metrics (like accuracy and training time).
In this example the training, validation and holdout rating entries are randomly sampled non-zero entries from the full ratings matrix $A$, summarized in the diagram below:

\vspace{-3mm}
\begin{figure}[H]
	\centering
	\includegraphics[width=\linewidth]{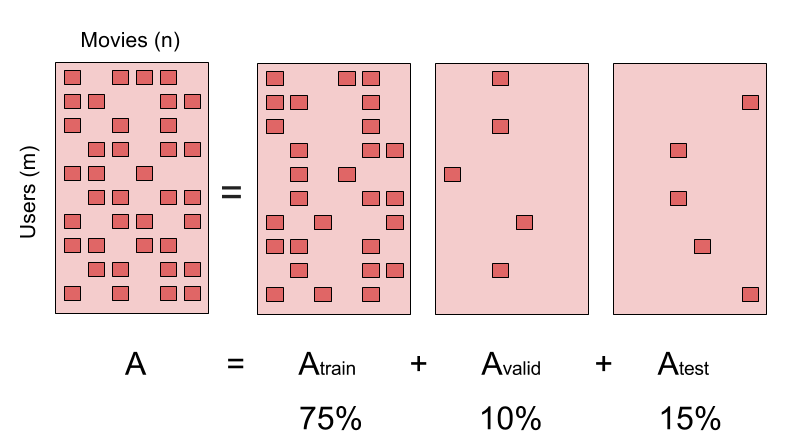}
	\caption{Train, validation and test sets for user movie ratings prediction}
\end{figure}

SigOpt tunes the alternating least square algorithm parameters with respect to the root mean squared error of the validation set.  We also report the performance on the hold out set as a measure of how well the algorithm generalizes to data it has not seen.  We compare parameters tuned using SigOpt against leaving the alternating least square parameters untuned. While the ratings entries for the train, valid and test sets were randomly sampled, they were identical sets in the SigOpt and the untuned comparisons.

\bgroup
\def\arraystretch{1.6}
\begin{table}[H]
\begin{center}
	\begin{tabular}{ |>{\centering}m{1.5cm}|>{\centering}m{2.5cm} |>{\centering}m{2cm} |>{\centering}m{3.5cm}| } 
		\hline
		 & SigOpt & Random Search &  \hspace{3mm}  No Tuning   \newline (Default MLlib ALS) \tabularnewline
		\hline
		Hold out \newline RMSE & \bf{0.7864} ({\color{ForestGreen}{-40.7\%}}) & 0.7901 & 1.3263 \tabularnewline
		\hline
	\end{tabular}
	\vspace{4mm}
	\caption{Comparison of RMSE on the hold out (test) ratings after tuning ALS algorithm}
\end{center}
\end{table}
\egroup
\small

\small
\bibliographystyle{plain}
\bibliography{citations}

\end{document}